\documentclass[5p,times,procedia]{elsarticle}
\usepackage{ecrc}
\volume{00}
\firstpage{1}
\journalname{Procedia Manufacturing}
\runauth{Deshpande et al.}
\jid{promfg}
\usepackage{amssymb}
\usepackage[figuresright]{rotating}

\usepackage{hyperref}
\hypersetup{colorlinks,allcolors=black}

\usepackage{subcaption}
\usepackage{url}

\begin{document}
\begin{frontmatter}



\dochead{48th SME North American Manufacturing Research Conference, NAMRC 48, Ohio, USA}%
\title{Computer Vision Toolkit for Non-invasive Monitoring of Factory Floor Artifacts}

\author[a,c]{Aditya M. Deshpande \corref{cor1}}
\author[b,c]{Anil Kumar Telikicherla \corref{cor1}}
\author[b,c]{Vinay Jakkali \corref{cor1}}
\author[d]{David A. Wickelhaus}
\author[a,c]{\newline Manish Kumar \corref{cor1}}
\author[b,c]{Sam Anand \corref{cor1}\fnref{correspondence_1}}

\address[a]{Cooperative Distributed Systems Lab}
\address[b]{Center for Global Design and Manufacturing}
\address[c]{Department of Mechanical and Materials Engineering\\University of Cincinnati, Cincinnati, Ohio 45221, USA}
\address[d]{TechSolve, Inc., Cincinnati, Ohio, USA}

\cortext[cor1]{Email: \{deshpaad, telikiav, jakkalvy\}@mail.uc.edu, \{sam.anand, manish.kumar\}@uc.edu
\newline Email addresses are given in order of author names.}
\fntext[correspondence_1]{Corresponding author.}

\begin{abstract}
Digitization has led to smart, connected technologies be an integral part of businesses, governments and communities. For manufacturing digitization, there has been active research and development with a focus on Cloud Manufacturing (CM) and the Industrial Internet of Things (IIoT). This work presents a computer vision toolkit (CV Toolkit) for non-invasive digitization of the factory floor in line with Industry 4.0 requirements for factory data collection. Currently, technical challenges persist towards digitization of legacy systems due to the limitation for changes in their design and sensors. This novel toolkit is developed to facilitate easy integration of legacy production machinery and factory floor artifacts with the digital and smart manufacturing environment with no requirement of any physical changes in the machines. The system developed is modular, and allows real-time monitoring of production machinery. Modularity aspect allows the incorporation of new software applications in the current framework of CV Toolkit. To allow connectivity of this toolkit with manufacturing floors in a simple, deployable and cost-effective manner, the toolkit is integrated with a known manufacturing data standard, MTConnect, to “translate” the digital inputs into data streams that can be read by commercial status tracking and reporting software solutions. The proposed toolkit is demonstrated using a mock-panel environment developed in house at the University of Cincinnati to highlight its usability.
\end{abstract}


\end{frontmatter}



\section{Introduction}
\label{intro}

\begin{figure*}[ht]
\centering
\includegraphics[width=\linewidth]{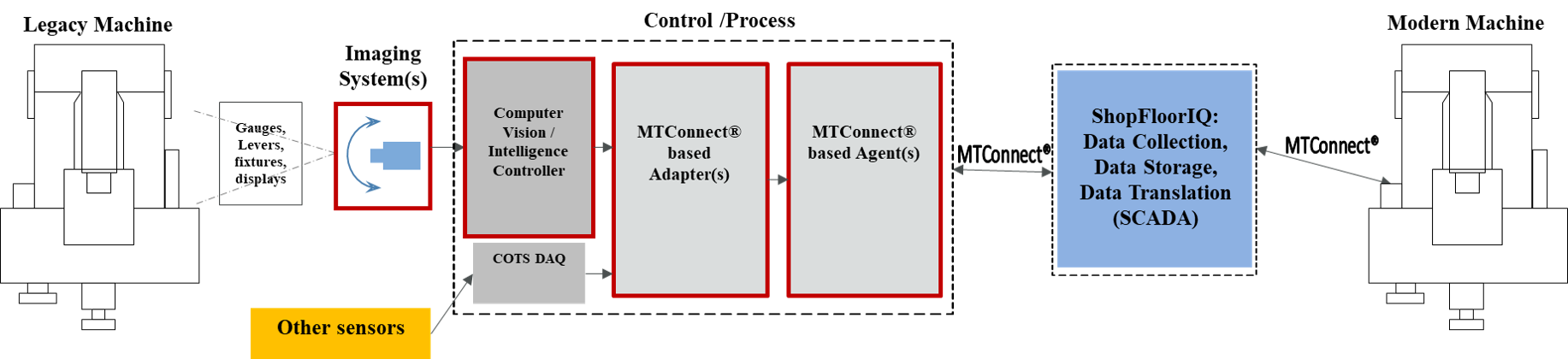}
\caption{Computer Vision Toolkit: This figure shows the schematic diagram of the CV Toolkit connected with the manufacturing floor; all the blocks with Red-colored border are the components of the toolkit; COTS DAQ stands for Commercial-Off-The-Shelf Data AcQuisition and SCADA is Supervisory Control And Data Acquisition. \label{system_schematic}}
\end{figure*}

With the dawn of Industry 4.0, manufacturing ecosystems are transforming intelligent and connected systems. This transformation has led to the automation of factory floors with the Industrial Internet of Things (IIoT), Cloud Manufacturing (CM) and Cyber-Physical Systems \cite{zhong2017intelligent, lee2015cyber, ren2015cloud}. This paradigm allows modern computerized machines to provide a wealth of data electronically, automatically and in near-real-time. This data provides insightful information to operations, maintenance, and management to quickly identify and resolve issues impacting productivity and quality.

Digitization has allowed for automation of manual tasks in manufacturing but the adaptation of these technologies in legacy manufacturing facilities has their own challenges \cite{rosas2017approach, sisinni2018industrial}. To recognize achievements in digital manufacturing we must address issues surrounding the difficulties in acquiring information across legacy, proprietary, and non-computerized machinery. Legacy and non-computerized machines provide little or no data electronically. For such equipment, acquiring process information requires manual monitoring or the installation of various sensors and data acquisition devices. This approach is oftentimes slow, expensive, invasive, and often provides limited data. As a result of these factors, the vital legacy equipment may be left unmonitored.

Industry 4.0 Digital Manufacturing will reach its full potential only when we can collect, process and utilize the data from all entities of a manufacturing floor \cite{sanchez2016enhancing}. In this process, the ability to collect and analyze data from legacy machines is of prime importance. Process optimization, part quality, on-the-floor productivity, and machine maintenance are some of the many advantages of collecting data and analyzing it.

To understand how to effectively implement the digital factory concept across all machines and process operations, a valuable source of information is the operators that observe, listen, use, and monitor the manufacturing equipment. They collect and process operation data based on the visual cues (dials, displays, part location, etc.), particular job settings, the operation sounds, and more. These cues form a wealth of information for the operators to control and maintain productivity and quality. If the equipment is not digitized, this awareness and knowledge live a finite and sheltered life within the operator’s memory. Digitization of these cues can provide automated feedback, work instruction verification, early detection of operational issues, preventive maintenance, and “big” data collection for downstream applications. This forms the motivation for the work presented in this paper. 

In this work, we attempt to address the issues and limitations of legacy machine digitization using a non-invasive approach. We present an innovative, affordable and non-invasive Computer Vision Toolkit (CV Toolkit) that combines computing and networking technologies and can connect with commercially available software analytics platforms. This toolkit is developed with a camera as the primary sensor to allow remote monitoring of the factory floor artifacts using computer vision techniques. The toolkit is designed to be inexpensive and work in real-time and digitizing data using this toolkit is simple, deployable and cost-effective. To allow interfacing of this toolkit with commercial status tracking and reporting software solutions, the data stream output is designed using the manufacturing data standard of MTConnect\textsuperscript{\textregistered}.

The rest of the paper is organized as follows: Section \ref{literature} provides a literature review about the factory floor monitoring and provides a brief review of various vision-based systems used in the modern manufacturing industry. Section \ref{overview_section} presents the system level overview of the CV Toolkit. Section \ref{attribute_section} explains the various features available in the presented toolkit and section \ref{sysreq_section} gives the details of software and hardware used in this work. Section \ref{demo_section} describes the demonstration of our toolkit on a mock-panel imitating various artifacts available on the factory floors as well as its demonstration when deployed on-site. Section \ref{conclusion} presents the conclusions and the future work.

\section{Literature Review}
\label{literature}

\begin{figure*}[thbp]
    \centering
    \begin{subfigure}{0.45\textwidth}
        \centering
        \includegraphics[width=\linewidth]{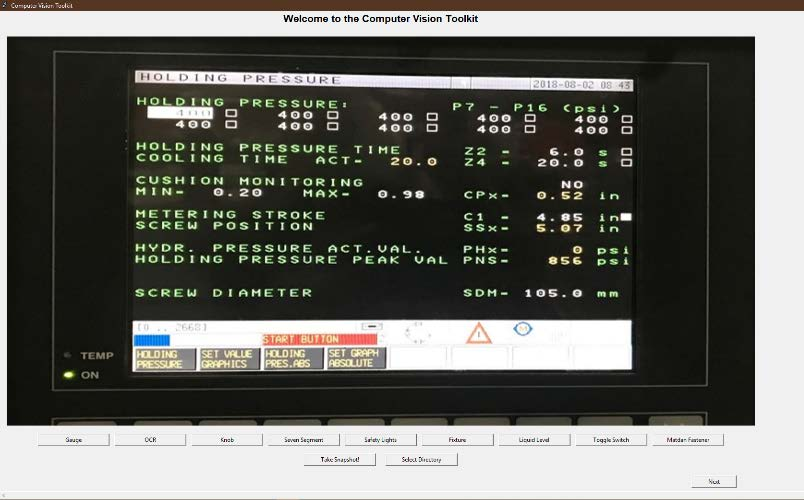}
        \caption{A camera feed of the LCD panel seen from the CV Toolkit. Below the camera feed, the GUI has the option for the user to select the appropriate application for the task of OCR. Image courtesy of Faurecia.}
    \end{subfigure}
    \hspace{3pt}
    \begin{subfigure}{0.45\textwidth}
        \centering
        \includegraphics[width=\linewidth]{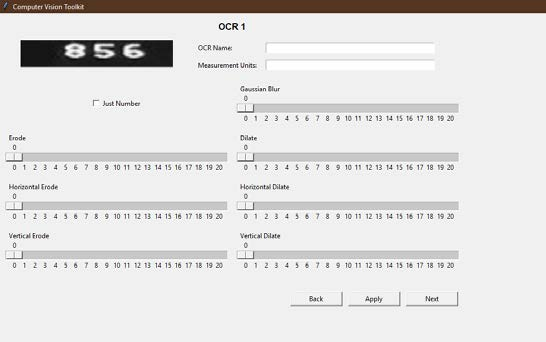}
        \caption{After selecting the ROI from the video feed, the user can navigate to the training phase of the application where it allows to set parameters required for image processing subroutines and MTConnect\textsuperscript{\textregistered} data stream. \label{gui_sample_ocr}}
    \end{subfigure}
    \caption{Graphical User Interface for CV Toolkit \label{gui_sample}}
\end{figure*}

Internet of Things (IoT) has enabled the development of powerful industrial applications with the use of ubiquitous technologies including wireless, mobile and sensor devices and internet-based information systems. As a result, several scientific studies have been conducted in this area in the industry as well as academia \cite{gershenfeld2004internet}. Various approaches have also been proposed to implement the Industry 4.0 in the current manufacturing infrastructure \cite{qin2016categorical}. 

Machine vision plays a major role in industrial IoT to optimize the processes for quality assurance. This is because of the low cost and ease of availability of high-quality data from vision sensors \cite{luckenhaus2016machine}. With the advancements in the areas of deep neural networks (DNNs), high compute and artificial intelligence, computer vision algorithms have achieved near-human accuracy in tasks such as object detection and localization \cite{zhao2019object, greenspan2016guest, szegedy2013deep}. Fast and reliable inspection of manufactured goods is now possible in modern manufacturing facilities as a result of this progress.

Modern factories have developed intending to resolve problems arising in production with flexible and adaptive processes \cite{radziwon2014smart}. The authors in \cite{yen2017framework} have proposed the Software-as-a-service (SaaS) framework for monitoring and diagnosis of systems in production at such facilities.
The work in \cite{rosas2017approach} proposed an approach to integrate the legacy systems with the IoT paradigm. It demonstrated a direct connection between legacy machine input/output circuits and the data distribution lines with appropriate PLCs. Research in \cite{tedeschi2017new} presented a cloud-based method to enable monitoring and data collection from legacy systems. Although this method elegantly addressed security challenges in IIoT-enabled legacy machinery monitoring, the authors focused mainly on  remote controllability of actuators on legacy machines.

Various vision-based systems for specific tasks have been proposed in the manufacturing industry. Product inspection and quality control are the major areas where these systems have proven to be useful. The work in \cite{martinez2019vision} has proposed the vision system for steel inspection. Authors in \cite{aminzadeh2019online} have developed a quality control framework for powder-bed additive manufacturing. In this work, a Bayesian classifier was trained using images captured from the high-resolution camera to identify the quality of metal powder-bed additive manufacturing processes and detect the defects in the manufactured product. A monocular-camera based system for autonomous forklift vehicles in a real factory environment is proposed in \cite{syu2017computer} for assisting in automated storage and retrieval tasks. IIoT and vision-systems have also aided in increasing sustainability in the manufacturing industry \cite{schluter2018vision}. Computer vision and mixed reality including augmented reality (AR) and virtual reality (VR) have enabled the enhancement of modern factory floors. A detailed review of AR, VR and computer vision in the manufacturing industry can be found in \cite{frigo2016augmented, damiani2018augmented, bottani2019augmented}. AR and vision systems have aided in increasing the efficiency of manual workstations \cite{uva2018evaluating}. VR has proven to be helpful on digital manufacturing floors for planning and conducting assembly operations \cite{abidi2019assessment}. The proposed CV toolkit provides a unified platform for various applications developed for assisting on the manufacturing floor using computer vision methods.

With digital factory concepts becoming reality, the need for connecting the product and process engineers to data from shop-floors/operations/equipment with reliable feedback for simulations, optimizations and controls is a priority. With consideration of a modern manufacturing facility with numerous systems operating in parallel, the seamless data transfer as a result of automation requires a standardized interface. MTConnect\textsuperscript{\textregistered} provides one such interface \cite{vijayaraghavan2008improving}. It is an open communication standard that enables data transfer on the manufacturing factory floor in an understandable format that can be read by any other device using the format. MTConnect has paved a way for unifying the data acquisition framework for factory floors in the age of Industry 4.0. Since its creation, MTConnect has found various applications on the manufacturing floor including but not limited to machine interoperability \cite{lei2016mtconnect}, system and quality monitoring \cite{edrington2014machine}, data handling in cyber-physical systems \cite{liu2019cyber}, cloud-based digital twin integration with manufacturing facilities \cite{hu2018modeling}. Since MTConnect is a widely accepted standard for ``seamless data-communication pipelines" in the manufacturing industry, it was adopted in the CV Toolkit system as a standard of communication.

\begin{figure}
    \centering
    \includegraphics[width=\linewidth]{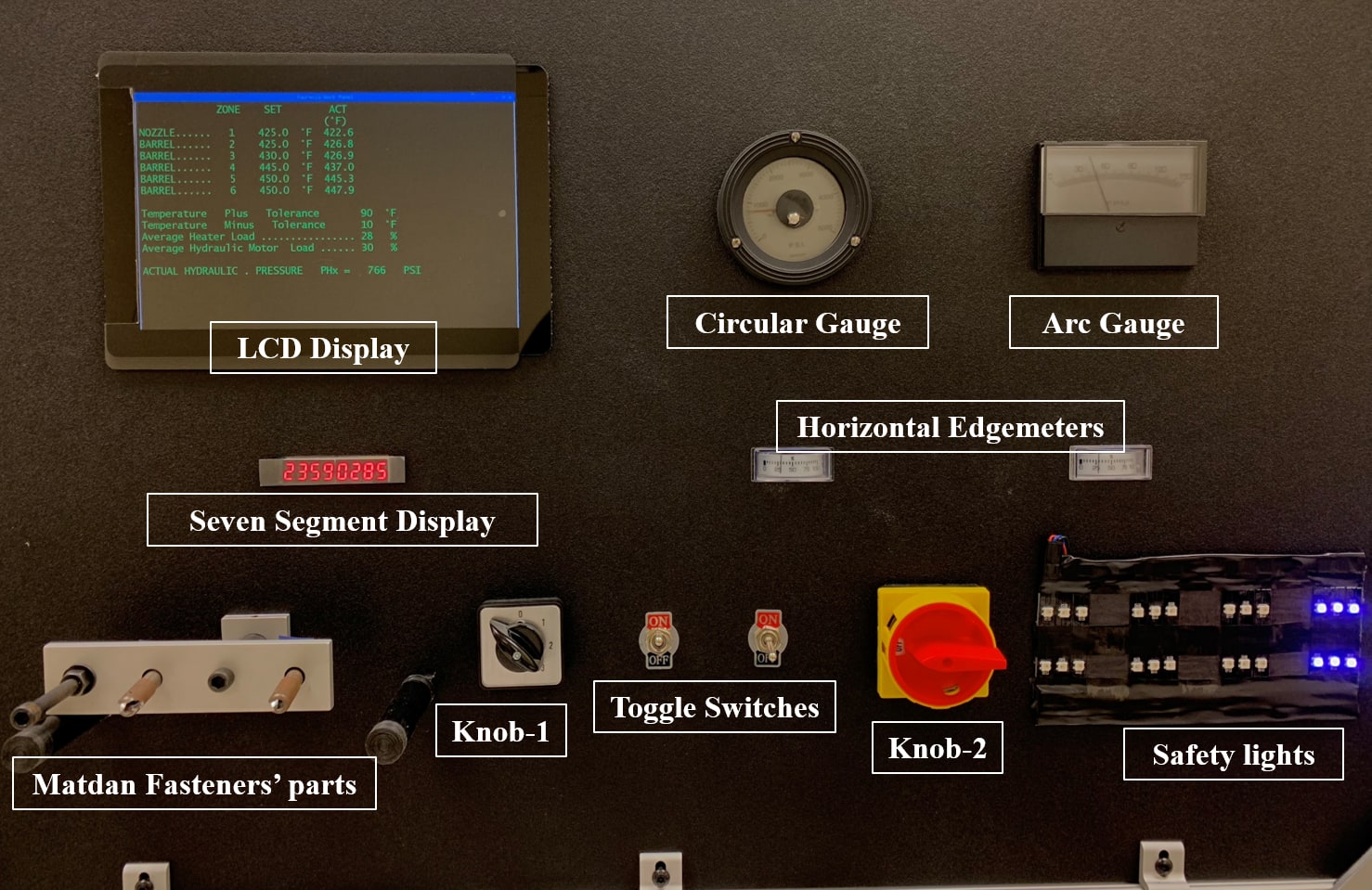}
    \caption{Mock panel built with various factory floor artifacts \label{mock_panel_image}}
\end{figure}

The use of DNNs and computer vision for fault detection in the printing industry with a context of Industry 4.0 was presented in \cite{villalba2019deep}. Although this application demonstrated immediate improvement in fault detection, there is a large scope of improvement which may include classification of fault categories as well as implementing this approach for monitoring the state of production machinery.
Solutions in IIoT space for manufacturing have mainly considered data-intensive predictive analysis tooling for automated maintenance of the factory floor and invasive solutions to automate the legacy systems \cite{ji2017bigdata, chen2019improving, jeschke2017industrial}.
Non-invasive solutions for the digitization of legacy machines have not received enough attention. In our work, we attempt to solve this problem using computer vision-based toolkit development for monitoring various artifacts on the factory floor. Our contribution is twofold. Firstly, we present a solution to monitor the legacy machine artifacts which allows capturing the analog data in digital format in real-time. Secondly, with the presented framework we not only demonstrate the applicability of our work in quality control and fault detection of the manufacturing output but also demonstrate the capability of our framework to interface with commercially available software platforms to allow data analytics and potential to be used for cloud manufacturing. This toolkit can be viewed as a manufacturing floor assistant which eventually alleviates manual stress on the operators.

\section{CV Toolkit Overview}
\label{overview_section}

Figure \ref{system_schematic} shows the overall schematic of the CV Toolkit with all its components. The imaging system (sensing module) forms the first component of the toolkit. This consists of a camera and the mounting apparatus to secure the camera with a focus on the legacy machine artifacts to capture the data of interest.
The second component is the software system that is further divided into three modules:
(i) The Computer Vision Software – The data collected by the imaging system is captured, processed and analyzed in this module. It forms the core of the toolkit responsible for the digitization of the legacy machine output; 
(ii) MTConnect\textsuperscript{\textregistered} based Adapter – The digital output of the Computer Vision Software is collected by this module in JSON format which contains the respective artifact information and its corresponding parameters. It is responsible for parsing of this JSON data and creating a stream of MTConnect\textsuperscript{\textregistered} standard data which can then be collected by an MTConnect\textsuperscript{\textregistered} based Agent;
(iii) MTConnect\textsuperscript{\textregistered} based Agent – This module stores the timestamped legacy machine artifact data in MTConnect\textsuperscript{\textregistered} standard format and acts as a server which can broadcast the data on demand through an HTTP connection.

The third component of the toolkit system is the analysis software, ShopFloorIQ \cite{itishopflooriq2013}, which is responsible for visualization, statistical and data analysis. It serves as a central dashboard for all the legacy machine data that enables the operator to monitor and maintain all the legacy machines on the plant floor from a single remote location.

\section{Features and Attributes}
\label{attribute_section}

\subsection{User Interface for Training the Computer Vision Software}
\label{cv_gui}
Graphical User Interface (GUI) is the preliminary component of the CV toolkit. It enables the user to train the software for a particular artifact of the legacy machine and corresponding factory environment. This allows the user to take into account the environmental variations including lighting conditions, regions of interest in the video data to monitor and number of artifacts to monitor at a given camera site or workstation. It is a one-time offline process performed by the operator to tune the system for the artifact under observation. Figure \ref{gui_sample} shows sample screenshots from the training GUI used for setting up the Optical Character Recognition (OCR) Application to read LCD panels on the manufacturing machines. GUI for training has the following purpose:
\begin{itemize}
    \item Selecting and focusing on the Region of Interest (ROI) of the required artifact from the camera feed.
    \item Correcting the alignment of the image to ensure that the artifact image is consistently horizontal.
    \item Pre-processing and denoising the image with the help of various image processing routines.
    \item To set up the parameters corresponding to MTConnect\textsuperscript{\textregistered} data streams.
\end{itemize}

With the training component in the GUI of the toolkit, the user can set various parameters required for image processing in CV Toolkit. These parameters include the kernel sizes for filtering operations, viz., blurring, contour extraction and median filtering \cite{buades2005review}. The other parameters the user can set include the data parsing elements for MTConnect\textsuperscript{\textregistered} client-service communication. These include application-specific parameters of the artifact being monitored, viz., the unique identifier for the MTConnect\textsuperscript{\textregistered} data stream, type of artifact, measurement units of the values being inferred in the application and time of measurement. The applications using template matching allow the user to select the set of ground truth templates based on the algorithm's specifications.

\begin{figure}
    \centering
    \includegraphics[width=\linewidth]{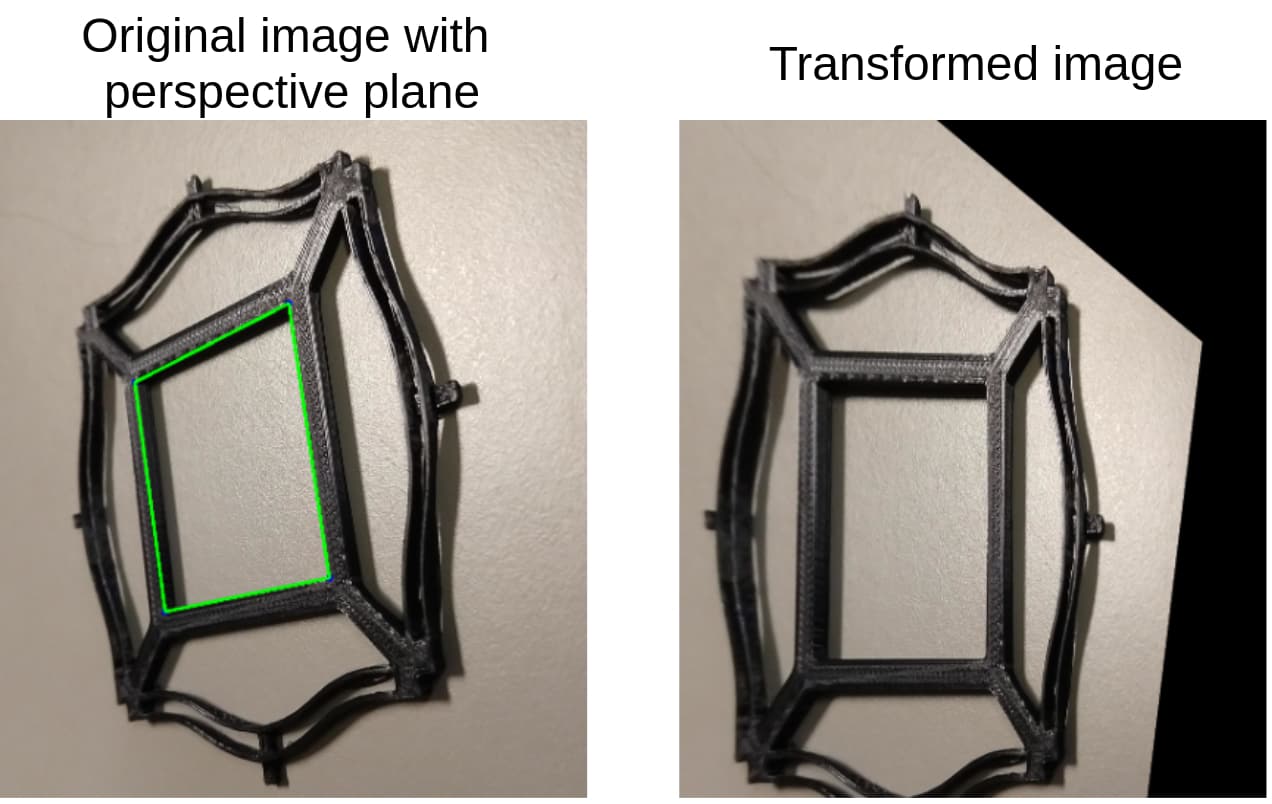}
    \caption{Illustration of perspective transform used to correct camera misalignment for artifacts. The left image was captured with a misaligned camera. The green contour marks the plane of interest. The transformed image can be seen on the right. (Note: The misalignment is magnified only for illustration purposes.) \label{perspective_transform}}
\end{figure}

All the monitoring applications in the toolkit are designed to look along the normal to the plane of the artifacts of interest. The camera misalignment was corrected using perspective transformation methods \cite{lehmann1999survey}. The transformation matrix was derived based on the points selected by the user from the plane associated with the artifact. The illustration of this transformation can be seen in fig \ref{perspective_transform}.

\subsection{Applications to interpret artifacts on the manufacturing floor}
\label{artifact}
This section briefly describes all the applications currently available in the CV Toolkit.

\subsubsection{Gauge Reader}
\label{gauge_app}
This application is designed to recognize the value of the linear and circular gauges which may be present on the legacy production machines. The algorithm employed in the toolkit is based on detecting the needle in the gauge. The GUI allows the image preprocessing similarly as illustrated in the case of OCR in fig. \ref{gui_sample_ocr}. This step allows the user to train the toolkit to segment out the needle from the ROI. In the inference phase, the application tracks the needle to estimate its position in the gauge.

\subsubsection*{Circular Gauge Reader}
This application is built to monitor the circular gauges. For interpreting the value of the gauge, the algorithm performs the linear mapping of the gauge scale to the angles of the needle. The minimum and the maximum of the gauge scale are taken as input from the user for the mapping. The needle is extracted from the image using the Hough line transform method\cite{fernandes2008real}.

\subsubsection*{Linear Gauge Reader}
A similar approach applying the Hough line transform was used for the extraction of the needle in linear gauges. The mapping is performed from the gauge scale to the needle position along horizontal or vertical as per the needle movement.

Equation \ref{interpolatoin_linear} is the linear interpolation used for mapping the needle reading to gauge scale. The needle reading is evaluated in radians for circular gauge and along X or Y axes depending upon the axis of needle movement of the linear gauge.

\begin{equation}
    \label{interpolatoin_linear}
    y_{out} = (y_{min} - y_{max}) \frac{x - x_{min}} {x_{max} - x_{min}} + y_{min}
\end{equation}

Here, $y$ represents the reading evaluated in the gauge scale and $x$ represents the needle reading evaluated from the algorithm which can be angular (in radians) or linear (pixel locations along X or Y axis) depending on the gauge type.

\subsubsection{OCR Applications}
\label{ocr_app}

\begin{figure}
    \centering
    \includegraphics[width=\linewidth]{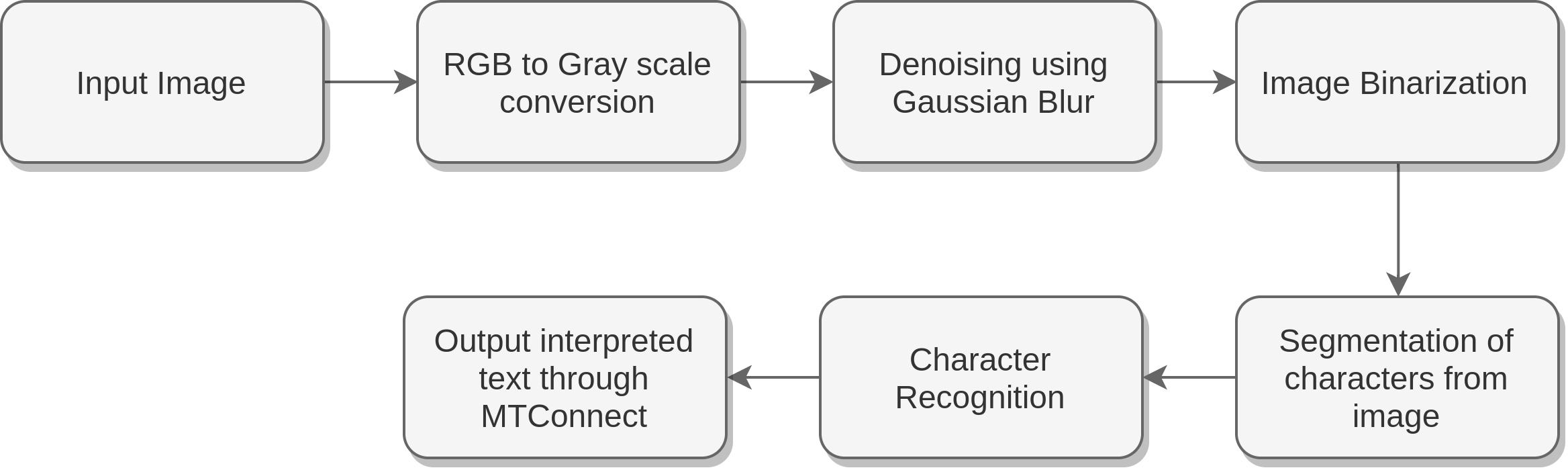}
    \caption{OCR application design flow \label{lcd_ocr_flow}}
\end{figure}

\begin{figure}
    \centering
    \includegraphics[height=0.4\linewidth]{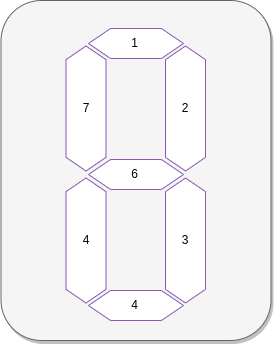}
    \caption{Seven Segment display template used in OCR \label{seven_segment_temp}}
\end{figure}

The optical character recognition module is built to recognize the text on displays available on the factory floor. There are two applications for the OCR which were developed in this work: (1) Seven segment display reader (2) LCD OCR to read the value of the character string displayed on the machine interface (LCD Screen). The application to read seven segment display was designed primarily to read the digits on a seven-segment digital display. These artifacts are usually available on manufacturing machine consoles to display the values of various environmental and process parameters. The OCR applications can be used to relay the machine information reliably to data-analytics platforms to analyze and identify various trends in machine parameters and process variables that may not be accessible from a computer. Figure \ref{lcd_ocr_flow} shows the outline of the OCR application flow. The character recognition block in this diagram performs the image to text mapping. The LCD OCR application was built using Tesseract OCR \cite{smith2007overview}. The seven-segment display reader is a simple template machining logic that uses a dictionary to match the set of lighted segment locations to a corresponding digit. The seven-segment display template used for this mapping can be seen in fig. \ref{seven_segment_temp}.

\subsubsection{Knob and Toggle Switch Monitors}
\label{knob_app}
These applications allow the user to track the states of various knobs and toggle switches on machine consoles. This application was developed to monitor the input states of machines. The state of toggle switches and knobs on the machines describe the input parameters. These states can be coupled with respective machine outputs to track and monitor overall system behavior.

\subsubsection{Machine Safety Lights Monitor}
\label{lights_app}
To detect the quality and safety conditions for the machines on the factory floor, safety lights with three colors such as red (unsafe), yellow (warning), green(safe) are installed. For legacy machines, these lights need to be continuously monitored by an operator on-site to take corrective actions in case of any malfunction or failures. To enable remote monitoring of the state of the legacy machine process, we have incorporated this application in the toolkit. This application uses the pixel intensities and RGB channel data to predict the color of the light being monitored. 

\subsubsection{Liquid Level Monitor}
\label{liquid_app}
Usually, the liquid level in chemical tanks in factory environments is monitored using intrusive sensors which are in direct contact with fluids \cite{reverter2007liquid}. In some cases, this may not be possible. To address this issue, we developed a vision-based liquid level monitoring application. The algorithm is based on the template machining \cite{briechle2001template, russell2013opencv} approach. The liquid level is marked by the user in the GUI which can be easily and efficiently tracked using a simple template matching approach based on the sum of the squared difference between the template and the image. This not only enables the detection of unsafe liquid levels, but tracking of exact liquid level is possible based on the base or zero references provided by the user.

\subsubsection{Fixture State Monitor}
\label{fixture_app}
Fixture State Monitor was developed primarily to detect the fixture orientations and patterns mounted on CNC machine beds. It was included as part of the CV Toolkit system to detect and identify any misaligned fixtures. During day-to-day changeover of parts on a machine, the fixtures can sometimes be clamped in a misaligned fashion which may lead to improper machining, scrap and machine downtime. With the help of the CV Toolkit, the operator can quickly identify if any of the fixtures are misaligned and make necessary corrections. The user needs to train the system by capturing template images of the fixtures in the correct orientation. The algorithm then compares individual fixtures in their surrounding area with the properly aligned fixture templates and identifies improperly aligned fixtures, if any.

\begin{figure*}
    \centering
    \begin{subfigure}{0.45\textwidth}
        \centering
        \includegraphics[width=\linewidth]{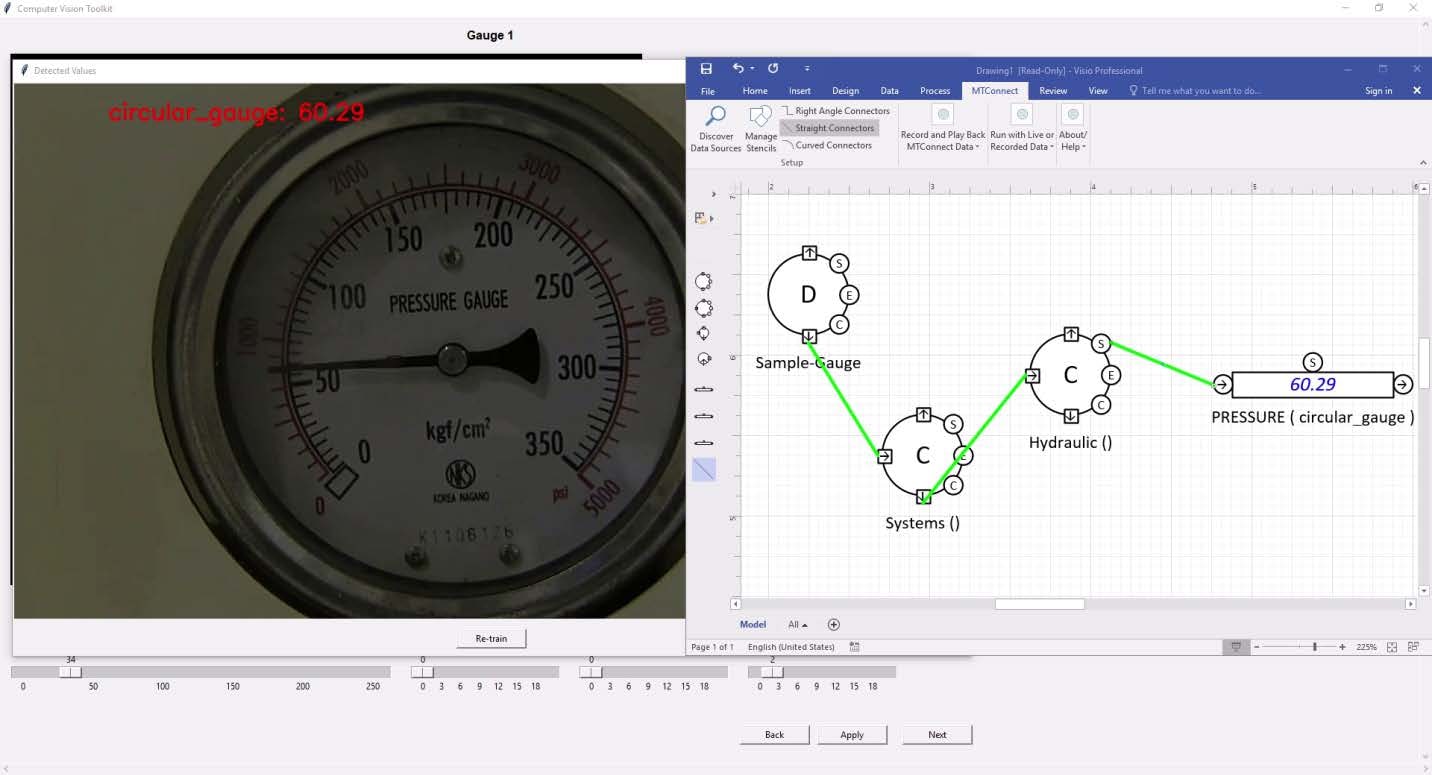}
        \caption{Gauge Reader interpreting the data from a circular gauge and the corresponding output is shown in ShopFloorIQ \label{circular_gauge_shopflooriq}. The data for this experiment is courtesy of Faurecia.}
    \end{subfigure}
    \hspace{3pt}
    \begin{subfigure}{0.45\textwidth}
        \centering
        \includegraphics[width=\linewidth]{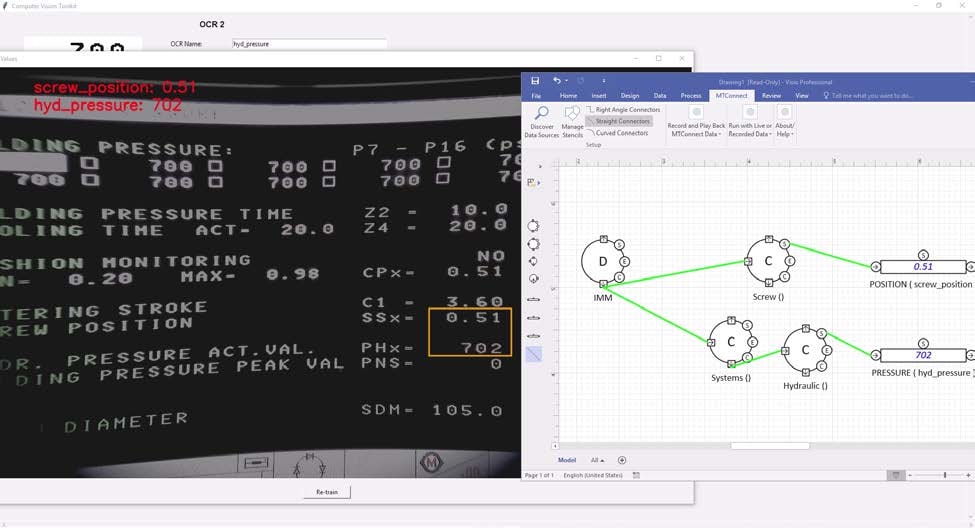}
        \caption{OCR Application reading data from the LCD display and showing the corresponding values in ShopFloorIQ \label{LCD_display_shopflooriq}. The data for this experiment is courtesy of Faurecia.}
    \end{subfigure}
    
    \begin{subfigure}{0.45\textwidth}
        \centering
        \includegraphics[width=\linewidth]{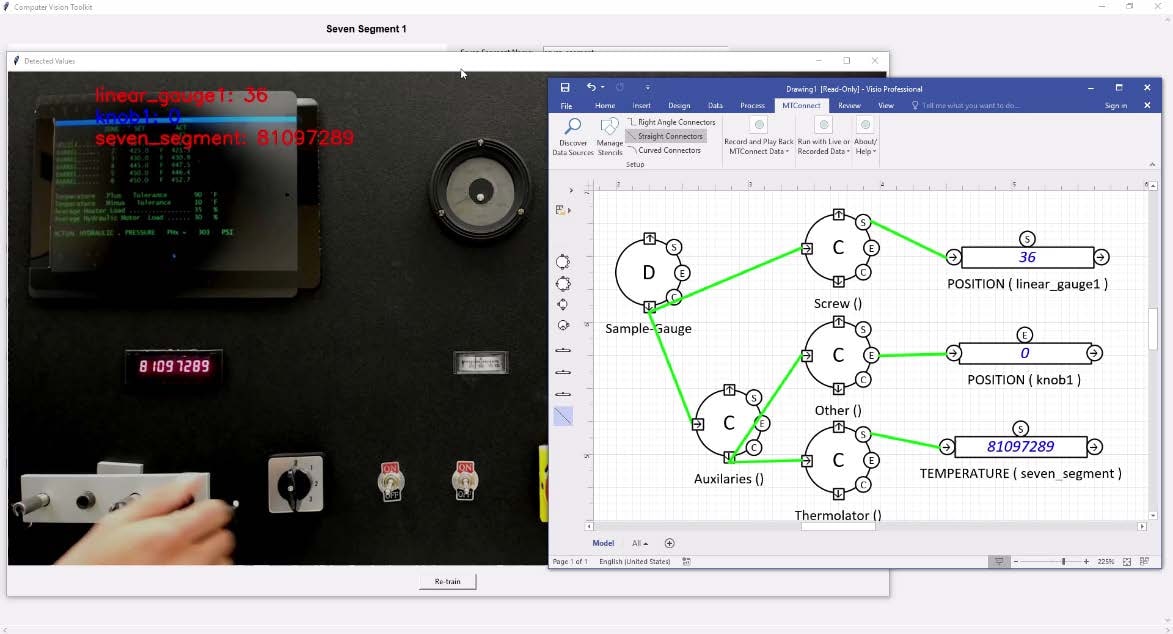}
        \caption{Knob and Toggle Switch Monitor application reading Knob-1, Horizontal Edge-meter/Linear gauge and Seven Segment Display (Refer fig. \ref{mock_panel_image}) and their corresponding outputs are shown in ShopFloorIQ. \label{knob_7seg_linear_gauge_shopflooriq}}
    \end{subfigure}
    \hspace{3pt}
    \begin{subfigure}{0.45\textwidth}
        \centering
        \includegraphics[width=\linewidth]{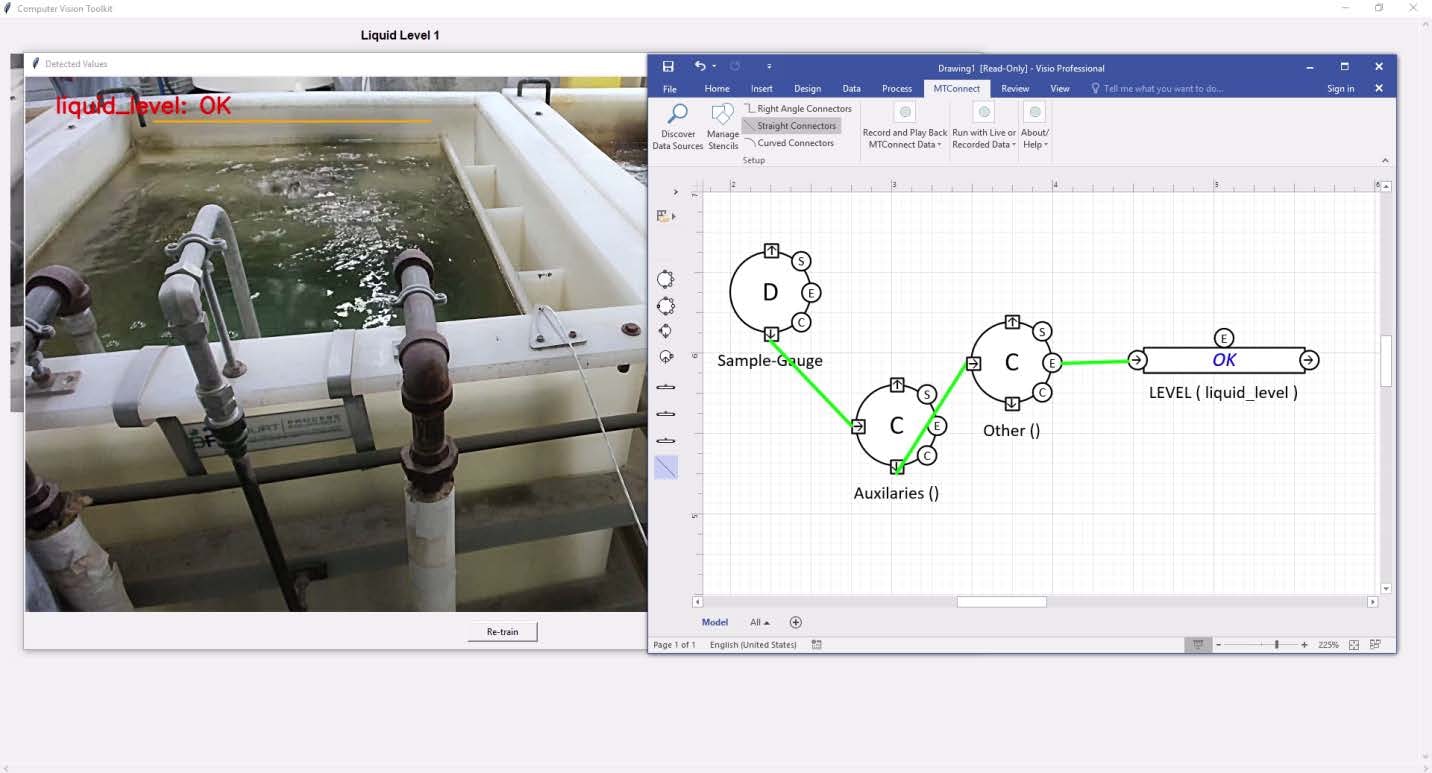}
        \caption{Liquid Level Monitor is shown in action with its ShopFloorIQ interface \label{liquid_level_shopflooriq}. The data for this experiment is courtesy of Raytheon Company.}
    \end{subfigure}

    \begin{subfigure}{0.45\textwidth}
        \centering
        \includegraphics[width=\linewidth]{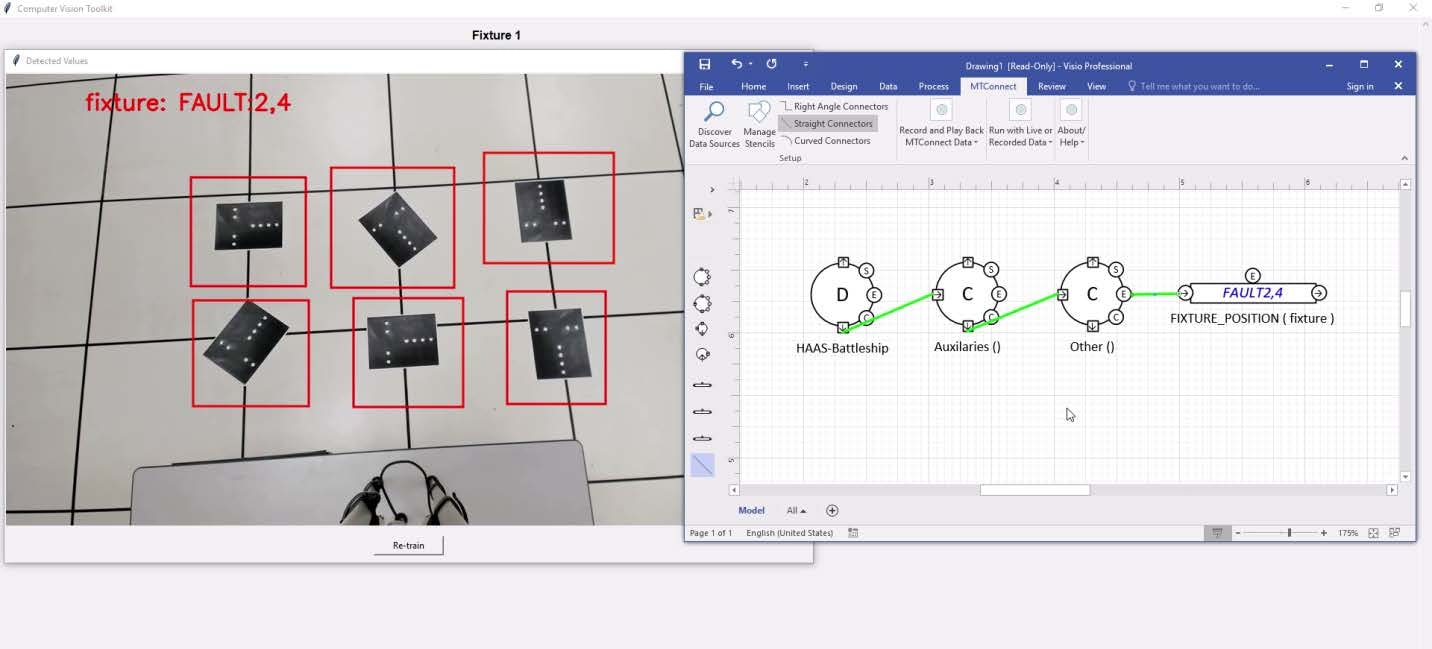}
        \caption{Fixture State Monitor is used to track state of a simulated machine bed with six fixtures with its output tracked in ShopFloorIQ. \label{fixture_shopflooriq}}
    \end{subfigure}
    \hspace{3pt}
    \begin{subfigure}{0.45\textwidth}
        \centering
        \includegraphics[width=\linewidth]{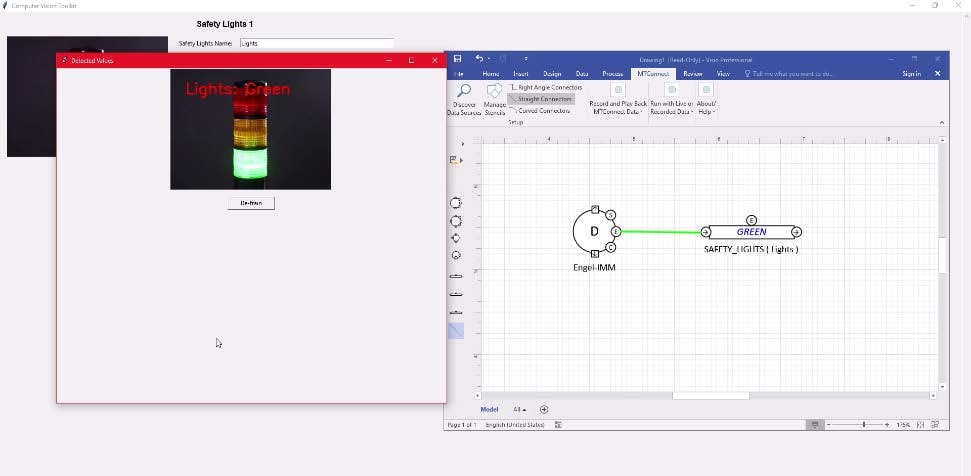}
        \caption{Machine Safety Lights Monitor tracking the state of safety lights and that state is sent to the ShopFloorIQ. \label{safety_lights_shopflooriq}}
    \end{subfigure}
    \caption{Demonstration of various CV Toolkit applications \label{artifact_demo_images}}
\end{figure*}

\subsubsection{Machine production quality Monitor}
\label{quality_app}
This application was developed to monitor the quality of the part produces on a CNC turning machine. The basis of this application is template matching based on the sum of squared difference \cite{briechle2001template} between the template of the good or properly machined part and the part received as the output of the machine. This makes it possible to detect the defective part output from the manufacturing line in real-time.

\subsection{Additional features of CV Toolkit}
\label{addition_features}
Following are some of the features of CV Toolkit which makes its adoption to the current manufacturing environment seamless:
\begin{itemize}
    \item Modular framework: Currently, this toolkit only contains the vision-based applications, but including other sensors is also possible to track non-visual states of the machines.
    \item Real-time and on-site processing: The video data from the cameras is processed on-site using the processing unit. The vision software with all the applications described in section \ref{artifact} is installed and run on-site.
    The video data is processed at the speed of 30fps. In fig. \ref{system_schematic}, this processing unit is referred to as Computer Vision/Intelligence Controller. The image transfer from on-site to remote location requires high bandwidth and can be computationally expensive \cite{chang2014bringing}. The on-site processing allows inferences on-site for review and transfer of processed data features to the remote locations. This makes it feasible for cloud-based applications.
    \item MTConnect standard integration: Various image processing and machine learning techniques are applied to the data that is captured by the imaging system and the digital output that is generated by the Computer Vision Software is broadcasted to MTConnect\textsuperscript{\textregistered} based Adapter through a TCP/IP network connection.
    \item External data analysis tool compatibility: The commercial data analysis tools such as ShopFloorIQ, can be used with this toolkit for further analysis, trends and preventive maintenance. These applications can be also used to trigger text/email alerts based on the digitized outputs from the toolkit. The data received from MTConnect\textsuperscript{\textregistered} based Adapter provides the flexibility to transmit and store data in cloud-based platforms allowing cloud-based computing and manufacturing.
\end{itemize}


\section{System Requirements}
\label{sysreq_section}
The CV Toolkit is developed using open-source software packages. The computer vision algorithms were developed using Python3 and OpenCV \cite{opencv_library}. The optical character recognition application was built using Tesseract OCR engine \cite{smith2007overview}. MTConnect\textsuperscript{\textregistered} standards \cite{mtconnect_standards} were used for data streaming over the network. To keep the cost requirement of the toolkit low and the viability to purchase off-the-shelf equipment along with easy installation, the hardware used for the prototype of CV Toolkit includes a Logitech BRIO 4K Pro Webcam as the visual sensor and an Intel NUC7i7BNH as the processing unit along with Ubuntu 16.04 LTS operating system in it. The computer vision software is platform-independent and it is compatible with edge computing units such as Raspberry Pi or NVIDIA Jetson TX2.

\section{Demonstration}
\label{demo_section}

This section presents the demonstration of CV Toolkit where it detects states of various factory floor artifacts described in section \ref{artifact}. Figure \ref{mock_panel_image} shows the mock-panel environment developed at the University of Cincinnati with the factory floor artifacts created to facilitate the development of these applications before deployment of CV Toolkit in an industrial environment. Currently, all the computer vision applications in the toolkit are interfaced with ShopFloorIQ \cite{itishopflooriq2013} through our MTConnect\textsuperscript{\textregistered} based Adapter and the figures show the state of the system along with the output received on the ShopFloorIQ software. Figure \ref{artifact_demo_images} shows the demonstration of these applications to detect state of various artifacts.

\begin{figure*}[thbp]
    \centering
    \begin{subfigure}{0.3\textwidth}
        \centering
        \includegraphics[width=\linewidth]{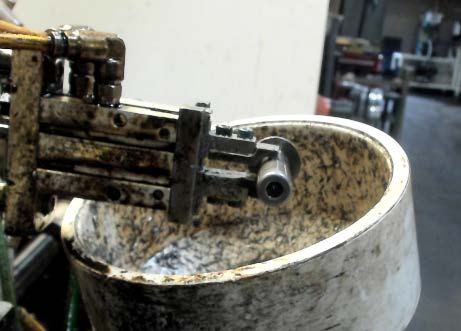}
        \caption{Unmachined part \label{quality_sample_1}}
    \end{subfigure}
    \hspace{3pt}
    \begin{subfigure}{0.3\textwidth}
        \centering
        \includegraphics[width=\linewidth]{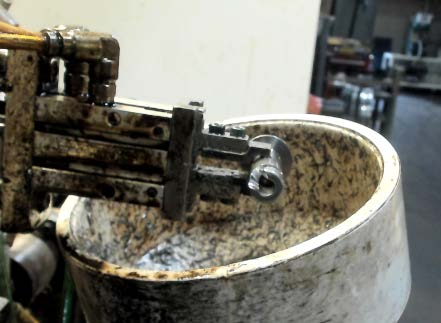}
        \caption{Improperly machined part \label{quality_sample_2}}
    \end{subfigure}
    \hspace{3pt}
    \begin{subfigure}{0.3\textwidth}
        \centering
        \includegraphics[width=\linewidth]{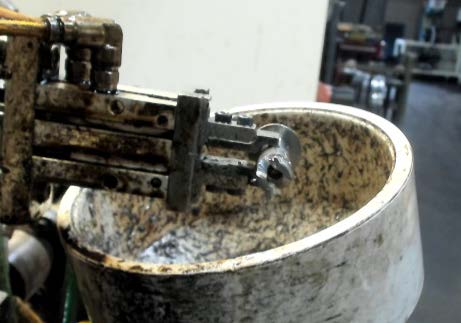}
        \caption{Properly machined part\label{quality_sample_3}}
    \end{subfigure}
    
    \begin{subfigure}{0.6\textwidth}
        \centering
        \includegraphics[width=\linewidth]{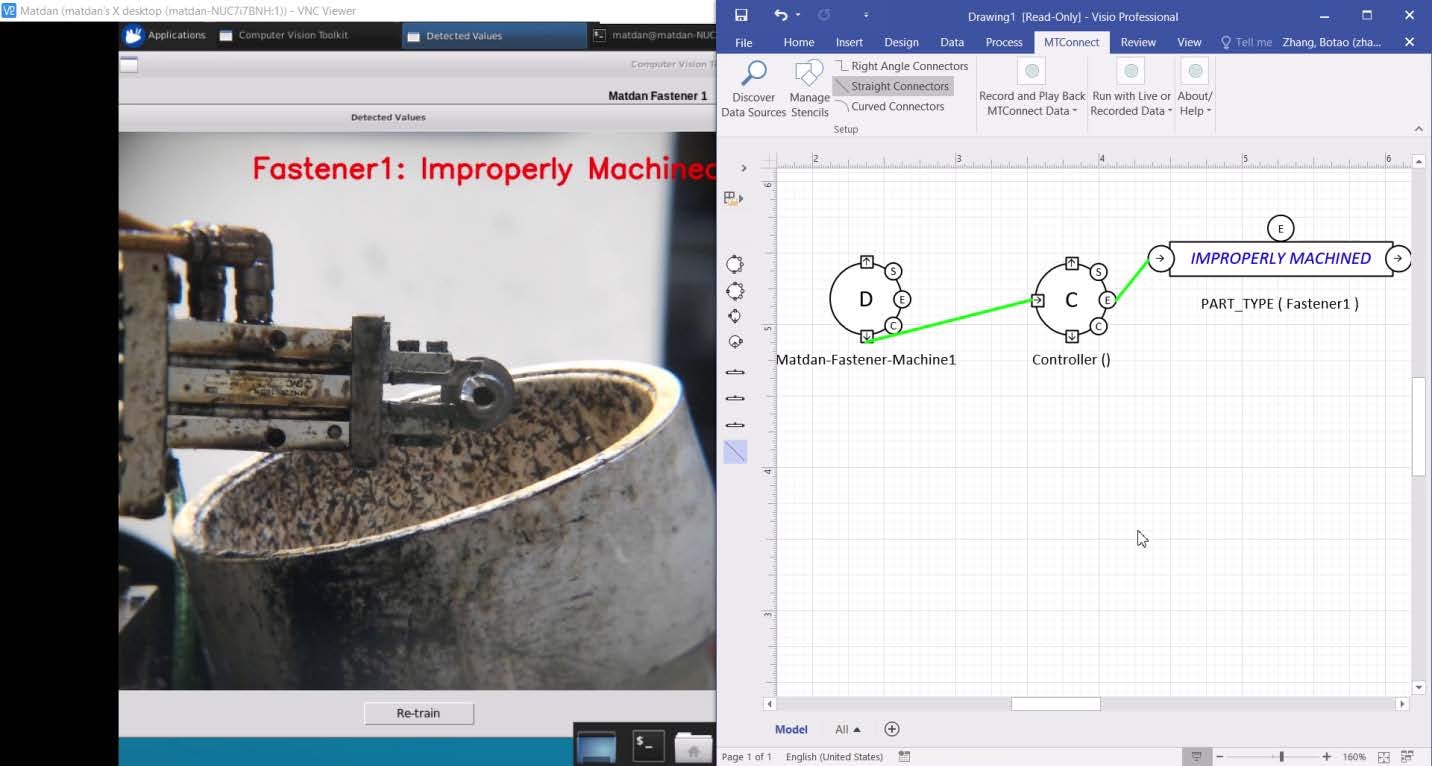}
        \caption{Machine production quality monitor with ShopFloorIQ \label{quality_demo_2}}
    \end{subfigure}
    \caption{Machine production quality monitor for defect detection of machine output. The data for this experiment is courtesy of Matdan Fastners. \label{quality_shopflooriq}}
\end{figure*}

Figure \ref{quality_shopflooriq} presents the use of this application for monitoring the quality. In this figure, the \emph{machine production quality monitor} application from CV Toolkit is used to track the output of the fastener manufacturing machine at Matdan\textsuperscript{\textregistered} Fasteners production facility. The figures \ref{quality_sample_1}, \ref{quality_sample_2} and \ref{quality_sample_3} show the possible outputs from the machine. Fig. \ref{quality_sample_1} shows the unmachined part coming out from the manufacturing machine. This figure indicates the part prior to the machining operation on the machine tool. Fig. \ref{quality_sample_2} indicates the partially machined part which indicates that the part is incorrectly processed by the machine tool and the machine requires replacement of the tools. The desired correct output of the fastener is shown in fig. \ref{quality_sample_3}. The image in this figure indicates that the machine is functioning correctly.
Figure \ref{quality_demo_2} shows the quality monitor application in the CV Toolkit in action with its output tracked on ShopFloorIQ. The output shown on the left of this figure corresponds to the case shown in fig. \ref{quality_sample_2}. The right side of fig. \ref{quality_demo_2} shows the corresponding digitized output received on the ShopFloorIQ GUI through MTConnect\textsuperscript{\textregistered} adapter-agent interface.

For the artifacts with reflective surfaces such as gauges (shown in fig. \ref{circular_gauge_shopflooriq}), it was observed that the sensing and inference of the developed applications were highly dependent on the light conditions and image resolution. A direct reflection of the light source on these artifacts may cause misinterpretation of the images because of the glare. To handle this noise, the artifacts were equipped with anti-glare screens while installing the toolkit for monitoring those artifacts.
The contribution of the automation of the inspection processes shown in figures \ref{artifact_demo_images} and \ref{quality_shopflooriq} is two fold. Firstly, this inspection process aids in reducing the manual fatigue while working with the artifacts demonstrated in these figures. Secondly, it enables data logging and digitization for various inspection processes in the manufacturing facilities. The data gathered can be used with the commercially available tools such as ShopFloorIQ for visualization and predictive maintenance of the machines. Another advantage of this system is the feasibility to install the cameras at locations where it may be difficult for a worker to reach. One such example is to monitor an artifact installed at a height such as a gauge that may be installed on a pipe carrying fluids through various levels of the manufacturing facility.

\section{Conclusion and Future work}
\label{conclusion}
This work developed a ``plug and play" toolkit framework to support a multitude of operations on the factory floor and produce information in the accepted and growing MTConnect\textsuperscript{\textregistered} standards. The developed CV Toolkit has been designed to acquire and digitize data from various legacy machine components (gauges, readouts, dial positions, and other shop floor artifacts). This system is non-invasive, i.e., there is no requirement for physical connections and/or internal modifications in the legacy machines. Leveraging MTConnect\textsuperscript{\textregistered}, this toolkit provides an intuitive graphical user interface to enable information collection and has complete compatibility with existing MTConnect\textsuperscript{\textregistered} applications. The authors envision that this technology will provide tremendous value for Original Equipment Manufacturers (OEM) as well as small and medium-sized enterprises (SME) to troubleshoot shop floor problems on the fly, reduce scrap rates, as well as perform process control adjustments based on real data and trend analysis of historical data.

The presented version of the CV Toolkit is developed with some constraints on the camera placement. The disturbance in the camera position requires the complete manual reset of the toolkit. This feature requires the development of an automated reset. To incorporate this automation and make the toolkit robust to external disturbances, addition of self-adapting features for interpretation and rejection of disturbance can a part of future work. As with any vision system, this toolkit is sensitive to light conditions of the environment and the parameters for all the applications are sensitive to light conditions. To address this limitation and make this system light condition agnostic, future work also includes improving the current image processing routines for low light conditions \cite{chen2018learning}. The current design of the toolkit was undertaken to keep it low cost. To make this system more robust and deployable even in hazardous environments such as the chemical plants, the toolkit design can be improved with appropriate enclosures on the various parts of the toolkit.

Future expansion of the toolkit could provide integration of additional sensor technology such as acoustic, vibration, and high-speed controller data developed through the community. The computer vision toolkit is non-invasive to the machine and operators and affordably provides the intelligence that the US industry needs to gain insight and make decisions from legacy and non-computerized equipment. Further additions of artificially intelligent modules using deep learning methods \cite{nisa2019critical} to detect and localize variations in the state of machines can be a good addition to the CV Toolkit.

\section*{Acknowledgements}
We thank TechSolve, Inc. for their support with the work on MTConnect standards and hardware setup for the mock-panel environment for testing. We would like to thank Kris Hill and International TechneGroup Inc. for their support while setting up the ShopFloorIQ environment. We thank Kristen Stone of Raytheon and Scott Bell of Faurecia for their valuable feedback in developing the CV Toolkit. We would also like to extend our gratitude to Raytheon, Faurecia and Matdan Fasteners for the support in deploying the CV Toolkit at their manufacturing facilities as part of the DMDII project. This work was supported by DMDII contract 16-02-06.

\bibliographystyle{vancouver}
\bibliography{reference_db}

\end{document}